\documentclass[]{spie}  

\usepackage{amssymb}
\usepackage{pifont}
\usepackage{adjustbox}

\usepackage{amsmath,amsfonts,amssymb}
\usepackage{subcaption}
\usepackage{graphicx}
\usepackage{cite} 
\usepackage{times}
\usepackage{epsfig}
\usepackage{amsmath}
\usepackage{nccmath}
\usepackage{amssymb}
\usepackage{mwe}
\usepackage{acro}
\usepackage{amssymb}
\usepackage{xcolor,colortbl}
\usepackage{tabularx}
\usepackage{relsize}
\usepackage{pifont}
\usepackage{booktabs} 
\usepackage{multirow}
\usepackage{multicol}
\usepackage{adjustbox}
\usepackage{float}
\usepackage{graphicx}
\usepackage{makecell}
\usepackage{tabu}
\usepackage[colorlinks=true, allcolors=blue]{hyperref}
\usepackage[capitalize]{cleveref}
\usepackage{soul, color, xcolor}

\title{M$^3$-GloDets: Multi-Region and Multi-Scale Analysis of Fine-Grained Diseased Glomerular Detection}

\author[a]{Tianyu Shi}
\author[b]{Xinzi He}
\author[c]{Kenji Ikemura}
\author[b,c]{Mert R. Sabuncu}
\author[c,d]{Yihe Yang}
\author[c]{Ruining Deng}

\affil[a]{Sichuan University, Chengdu, 610207, CN}
\affil[b]{Cornell Tech, New York, NY 10044, USA}
\affil[c]{Weill Cornell Medicine, New York, NY 10065, USA}
\affil[d]{Northwell Health, New Hyde Park, NY 11040, USA}

\begin{document} 
\maketitle

\begin{abstract} 
Accurate detection of diseased glomeruli is fundamental to progress in renal pathology and underpins the delivery of reliable clinical diagnoses. Although recent advances in computer vision have produced increasingly sophisticated detection algorithms, the majority of research efforts have focused on normal glomeruli or instances of global sclerosis, leaving the wider spectrum of diseased glomerular subtypes comparatively understudied. This disparity is not without consequence; the nuanced and highly variable morphological characteristics that define these disease variants frequently elude even the most advanced computational models. Moreover, ongoing debate surrounds the choice of optimal imaging magnifications and region-of-view dimensions for fine-grained glomerular analysis, adding further complexity to the pursuit of accurate classification and robust segmentation.

To bridge these gaps, we present M$^3$-GloDet, a systematic framework designed to enable thorough evaluation of detection models across a broad continuum of regions, scales, and classes. Within this framework, we evaluate both long-standing benchmark architectures and recently introduced state-of-the-art models that have achieved notable performance, using an experimental design that reflects the diversity of region-of-interest sizes and imaging resolutions encountered in routine digital renal pathology. As the results, we found that intermediate patch sizes offered the best balance between context and efficiency. Additionally, moderate magnifications enhanced generalization by reducing overfitting. Through systematic comparison of these approaches on a multi-class diseased glomerular dataset, our aim is to advance the understanding of model strengths and limitations, and to offer actionable insights for the refinement of automated detection strategies and clinical workflows in the digital pathology domain.

\end{abstract}


\section{Description of purpose}  
\label{sec:intro}  
Accurately detecting diseased glomeruli is central to renal pathology and essential for delivering precise clinical diagnoses. Despite significant advances in computer vision and object detection, most prior research has concentrated on the identification of normal glomeruli~\cite{gallego2018glomerulus, hao2023accurate} and cases of global sclerosis~\cite{altini2020deep}. In contrast, the detection and classification of fine-grained diseased glomerular subtypes, including global sclerotic, viable, ischemic, segmental sclerotic, and atubular forms, have not received comparable attention~\cite{zeng2020identification}. The distinctive and often subtle morphological features that define these disease variants present unique challenges, exposing the limitations of both traditional detection models and recently proposed state-of-the-art approaches. Consequently, the ability of state-of-the-art architectures to reliably distinguish and separate these complex pathological entities remains largely unexplored. 
This challenge is compounded by the lack of consensus on optimal imaging magnification~\cite{nan2022automatic} and region-of-view size~\cite{zhu2025cross} for fine-grained glomerular analysis. Effective classification depends on the model’s capacity to resolve intricate morphological details, while robust instance segmentation demands precise boundary delineation. The absence of standardized imaging protocols complicates both model development and objective assessment of automated detection systems, highlighting the importance of systematic, multi-scale, and multi-region evaluation. 

Several prior studies have reported encouraging results on the classification of diseased glomeruli~\cite{he2024image,lu2021improve}. However, the majority of these efforts have been limited to semantic segmentation of normal glomeruli~\cite{deng2023omni,deng2024prpseg,deng2024hats,deng2025segment,wang2025glo} or diseased lesions~\cite{yao2022glo,yu2025glo}, without extending to an instance-level segmentation framework. To date, only a few dedicated pipelines~\cite{jiang2021deep} have been proposed for the instance segmentation of diseased glomeruli, leaving a critical methodological gap in the field.  

To address these challenges, we introduce M$^3$-GloDet, a systematic framework designed to enable comprehensive evaluation of detection models across a continuum of spatial scales and region sizes for fine-grained diseased classes. As shown in Table~\ref{tab:method}, our investigation spans classic baseline architectures, including Mask R-CNN~\cite{he2017mask}, Cascade R-CNN~\cite{8917599}, and SOLOv2~\cite{wang2020solov2}, as well as recently proposed state-of-the-art models such as DiffusionInst~\cite{10447191}, ASF-YOLO~\cite{kang2024asf}, CelloType~\cite{pang2025cellotype}, and YOLOv12~\cite{tian2025yolov12}. Our experimental design accommodates a wide spectrum of region-of-interest sizes (2048$\times$2048, 4096$\times$4096, and 8192$\times$8192 pixels at 20$\times$ magnification) and imaging resolutions (20$\times$, 10$\times$, and 5$\times$ at 4096$\times$4096 pixels), capturing the inherent variability found in digital renal pathology practice. By systematically benchmarking these models on a multi-class dataset of diseased glomeruli, this work seeks to deepen understanding of model capabilities and limitations, while offering actionable guidance for future algorithm development, experimental design, and clinical translation in the evolving field of digital renal pathology.

\begin{table}[ht]
\centering
\begin{adjustbox}{width=\textwidth}
\renewcommand{\arraystretch}{1.1}
\begin{tabular}{lllcc l}
\toprule
\textbf{Method} & \textbf{Backbone} & \textbf{Year} & \textbf{\makecell{Params\\ (Billion)}} & \textbf{\makecell{Peak GPU\\Memory (GB)}} & \multicolumn{1}{l}{\textbf{Code Repository}} \\
\midrule
Mask R-CNN~\cite{he2017mask} & ResNet-50-FPN & 2017 & 0.044 & 13.9 & \url{https://github.com/facebookresearch/detectron2} \\
Cascade R-CNN~\cite{8917599} & ResNet-50-FPN & 2019 & 0.072 & 15.0 & \url{https://github.com/zhaoweicai/Detectron-Cascade-RCNN} \\
SOLOv2~\cite{wang2020solov2} & ResNet-50-FPN & 2020 & 0.047 & 12.3 & \url{https://github.com/aim-uofa/AdelaiDet} \\
DiffusionInst~\cite{10447191} &  Swin-Base & 2024 & 0.175 & 34.0 & \url{https://github.com/chenhaoxing/DiffusionInst} \\
ASF-YOLO~\cite{kang2024asf} & CSPDarknet53 & 2024 & 0.048 & 41.0 & \url{https://github.com/mkang315/ASF-YOLO} \\
CelloType~\cite{pang2025cellotype} &Swin-Large & 2025 &0.223  &41.3  & \url{https://github.com/tanlabcode/CelloType} \\
YOLOv12~\cite{tian2025yolov12} &R-ELAN & 2025 &0.003  &44.7  & \url{https://github.com/sunsmarterjie/yolov12} \\
\bottomrule
\end{tabular}
\end{adjustbox}
\caption{Seven detection methods evaluated in this study, including model size, peak GPU memory usage during training, and corresponding code repositories.}
\label{tab:method}
\end{table}

\begin{figure}[h]
\begin{center}
\includegraphics[width=0.95\textwidth]{{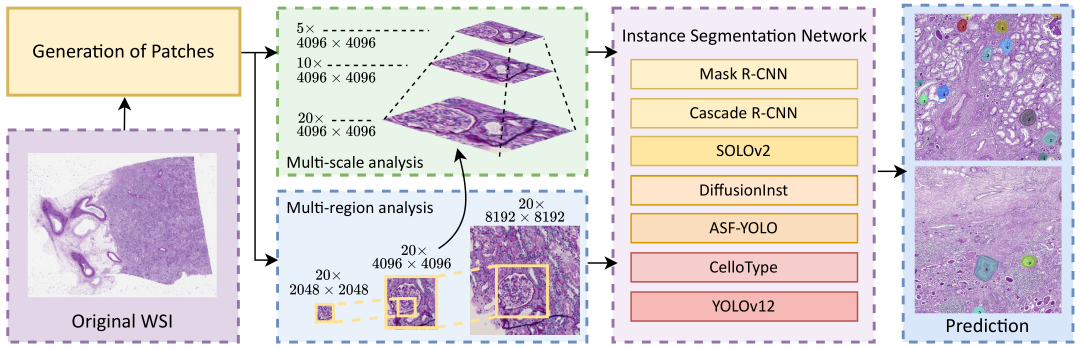}}
\end{center}
\caption{A pipeline of glomerular lesion-specific instance segmentation.} 
\label{fig1:Problem}
\end{figure}

\section{Method}  
\label{sec:method}

\subsection{Overall Framework}

Figure~\ref{fig1:Problem} presents an overview of the instance segmentation pipeline designed for whole-slide histopathology images (WSIs). To manage the large size and high resolution of WSIs, each slide is partitioned into smaller patches at three magnification levels: 5$\times$, 10$\times$, and 20$\times$. Patch sizes are set to 2048$\times$2048, 4096$\times$4096, or 8192$\times$8192 pixels, accommodating different levels of detail. During training, a patch sampler selects regions annotated as informative, purposefully omitting areas lacking glomeruli to reduce unnecessary computational load. Each selected patch is subsequently fed into an instance segmentation model, which produces object masks and their corresponding bounding boxes. These outputs form the basis for subsequent evaluation of segmentation and classification performance.

\subsection{Detection Models and Architectures}

We evaluate seven representative detection models spanning from foundational architectures to state-of-the-art approaches, arranged chronologically to illustrate the evolution of instance segmentation and object detection techniques:

\begin{itemize}
    \item[\textbf{1.}]\textbf{Mask R-CNN~\cite{he2017mask}:}  
    Mask R-CNN employs a ResNet-50-FPN backbone and is renowned for its two-stage detection framework. Its most innovative feature is the integration of a Region Proposal Network (RPN) with precise Region of Interest (RoI) alignment. This design enables the model to generate highly accurate object proposals across multiple scales and subsequently refine both localization and segmentation masks. The ability to finely localize and distinguish glomeruli, even when they are morphologically diverse or closely clustered, makes Mask R-CNN especially effective for detailed glomerular segmentation.

    \item[\textbf{2.}]\textbf{Cascade R-CNN~\cite{8917599}:}
    Cascade Mask R-CNN builds upon the ResNet-50-FPN backbone and extends the two-stage detection paradigm through its distinctive multi-stage refinement architecture. Its key innovation lies in the cascade of sequential detectors, where each stage is trained with progressively higher IoU thresholds to address the mismatch between training and inference conditions. This cascaded design enables the model to iteratively refine object proposals, with each subsequent stage operating on higher-quality detections from the previous stage. The progressive refinement mechanism proves particularly valuable for glomerular detection, where initial coarse proposals are systematically refined to achieve precise boundary delineation, especially beneficial for handling overlapping glomeruli and improving segmentation accuracy in challenging pathological scenarios.

    \item[\textbf{3.}]\textbf{SOLOv2~\cite{wang2020solov2}:}  
    SOLOv2 adopts a single-stage approach, featuring a ResNet-50-FPN backbone enhanced with deformable convolutions. What sets SOLOv2 apart is its innovative grid-based instance representation, which divides the input image into a structured grid and directly predicts the presence and segmentation mask for each grid cell. This proposal-free paradigm yields high computational efficiency and excels in scenarios where glomeruli are densely distributed and vary significantly in size. The incorporation of deformable convolutions further equips SOLOv2 to handle the irregular and complex contours that often characterize diseased glomeruli.

    \item[\textbf{4.}]\textbf{DiffusionInst~\cite{10447191}:} 
    DiffusionInst employs a Swin-Base backbone and introduces a novel paradigm by adapting diffusion frameworks to instance segmentation tasks. Its key innovation lies in representing instances as vectors and formulating segmentation as a noise-to-vector denoising process. Unlike conventional approaches, the model is trained to reverse noisy ground truth masks without relying on Region Proposal Network inductive biases. During inference, it generates instance masks through multi-step denoising from randomly initialized vectors, leveraging the iterative refinement capabilities inherent in diffusion models. For glomerular segmentation, DiffusionInst's vector-based representation and denoising approach offers advantages in handling complex morphological variations and overlapping structures, making it particularly suitable for distinguishing subtle pathological features across different glomerular types.

    \item[\textbf{5.}]\textbf{ASF-YOLO~\cite{kang2024asf}:}
    ASF-YOLO builds upon the YOLOv5 framework with a CSPDarknet53 backbone and introduces attentional scale sequence fusion for accurate and fast cell instance segmentation. Its key innovation lies in combining spatial and scale features through novel fusion modules that enhance multiscale information extraction and integrate detailed information from different scales. The model incorporates attention mechanisms to focus on informative channels and spatial positions, particularly improving performance for small object detection and segmentation. For glomerular segmentation tasks, ASF-YOLO's attentional fusion approach effectively handles the diverse morphological characteristics and scale variations of different glomerular types, while maintaining high inference speed suitable for practical pathological image analysis applications.

    \item[\textbf{6.}]\textbf{CelloType~\cite{pang2025cellotype}:} 
    CelloType utilizes a Swin Transformer backbone and introduces an end-to-end framework for joint cell segmentation and classification. Its key innovation lies in the multitask learning strategy that integrates segmentation and classification tasks simultaneously, leveraging the interconnected nature of these processes to enhance overall performance. The model combines three core modules: Swin Transformer-based feature extraction, DINO object detection and classification, and MaskDINO segmentation, all unified within a single neural network for end-to-end learning. CelloType employs deformable attention mechanisms and contrastive denoising training to improve detection accuracy and provides confidence scores for both segmentation and classification predictions. For glomerular segmentation tasks, CelloType's unified multitask approach offers potential advantages in handling complex morphological variations by simultaneously leveraging both spatial segmentation information and class-specific features, enabling more comprehensive analysis of diverse glomerular types within histopathological images.

    \item[\textbf{7.}]\textbf{YOLOv12~\cite{tian2025yolov12}:} 
    YOLOv12 builds upon a hierarchical backbone design and represents the successful integration of attention-centric architecture into the YOLO framework while maintaining real-time performance. Its key innovation lies in the Area Attention module, which reduces computational complexity through efficient feature map segmentation, and the Residual Efficient Layer Aggregation Networks (R-ELAN) that address optimization challenges in attention-based models. The model achieves superior accuracy-latency trade-offs by combining the global modeling capabilities of attention mechanisms with the computational efficiency required for real-time detection. For glomerular segmentation tasks, YOLOv12's attention-based design enables enhanced feature representation and improved contextual understanding, potentially offering advantages in handling complex morphological variations and spatial relationships within histopathological images.
    
\end{itemize}

\section{DATASET AND EXPERIMENTS} 

\subsection{Data} 
This study retrospectively included eight nephrectomy cases processed at the Department of Pathology, Northwell Health, comprising specimens from both Long Island Jewish Medical Center and Northshore University Hospital between July 2017 and October 2019. Whole-slide images (WSIs) were generated from de-identified, non-cancerous sections of kidney parenchyma stained with periodic acid–Schiff (PAS) and digitized using Leica GT450 RUO scanners at 20$\times$ magnification. Glomeruli within each WSI were initially identified in QuPath, after which expert renal pathologists performed detailed manual annotations. Each glomerulus was delineated with a pixel-level instance mask and categorized into one of five clinically significant groups: global glomerulosclerosis (GGS), viable glomerulus (Viable Glom), ischemic glomerulus (Ischemic Glom), segmental glomerulosclerosis (SGS), or atubular glomerulus (Atubular Glom). The total number of glomeruli in each category is presented in Table~\ref{tab:glomerular_counts}.

\begin{table}[ht]
\centering
\begin{adjustbox}{width=0.6\textwidth}
\renewcommand{\arraystretch}{1.2}
\begin{tabular}{cccccc}
\toprule
\textbf{Dataset Split} & \textbf{GGS} & \textbf{Viable Glom} & \textbf{Ischemic Glom} & \textbf{SGS} & \textbf{Atubular Glom} \\
\midrule
\textbf{Train} & 1338 & 520 & 211 & 233 & 79 \\
\textbf{Validation} & 154 & 18 & 51 & 17 & 15 \\
\textbf{Test} & 140 & 231 & 54 & 66 & 10 \\
\bottomrule
\end{tabular}
\end{adjustbox}
\caption{Distribution of annotated glomeruli by lesion subtype across training, validation, and test sets.}
\label{tab:glomerular_counts}
\end{table}

\subsection{Experimental Details} 
For this study, the dataset was partitioned into three distinct subsets: five slides comprised the training set, one was allocated for validation, and the remaining two were reserved for external testing. Each annotated glomerulus was assigned to one of the five previously defined lesion categories, facilitating fine-grained, multi-class segmentation and comprehensive pathological assessment.  All models were implemented using their official architectures and trained under consistent computational settings. All experiments were carried out on a dedicated workstation equipped with an NVIDIA A6000 Ada GPU, ensuring consistent computational performance throughout the analysis.

\subsection{Evaluation Metrics}

To comprehensively evaluate segmentation performance, we employed complementary metrics at both the instance and pixel levels. Specifically, macro-F1 was computed across the five glomerular categories, using an IoU threshold of 0.5. Mean Average Precision (mAP), calculated over IoU thresholds from 0.5 to 0.95 in increments of 0.05, provided a standardized measure of detection accuracy. Finally, the Dice coefficient quantified the spatial overlap between predicted and ground-truth masks, capturing the morphological fidelity of segmentation outcomes.

Dice scores were aggregated differently across analyses: for tabular results (Table~\ref{tab:performance_comparison}, Table~\ref{tab:multi_magnification_results}), we reported a single score per category using the full test dataset; for boxplots (Figure~\ref{fig3:boxplot}), scores were computed per patch, excluding patches without class instances. This approach ensured fair evaluation of rare categories and prevented bias from empty patches.

\newcommand{\up}{\ensuremath{\uparrow}}

\begin{table}[ht]
\centering
\begin{adjustbox}{width=\textwidth}
\renewcommand{\arraystretch}{1.2} 
\begin{tabular}{cc*{18}{c}}
\toprule
\multirow{2}{*}{\textbf{Method}} & \multirow{2}{*}{\textbf{Size}} &
\multicolumn{3}{c}{\textbf{GGS}} & \multicolumn{3}{c}{\textbf{Viable Glom}} & \multicolumn{3}{c}{\textbf{Ischemic Glom}} & 
\multicolumn{3}{c}{\textbf{SGS}} & \multicolumn{3}{c}{\textbf{Atubular Glom}} & \multicolumn{3}{c}{\textbf{Average}} \\
\cmidrule(lr){3-5} \cmidrule(lr){6-8} \cmidrule(lr){9-11}
\cmidrule(lr){12-14} \cmidrule(lr){15-17} \cmidrule(lr){18-20}
& & Dice\up & F1\up & mAP\up & Dice\up & F1\up & mAP\up & Dice\up & F1\up & mAP\up 
 & Dice\up & F1\up & mAP\up & Dice\up & F1\up & mAP\up & Dice\up & F1\up & mAP\up \\
\midrule
\multirow{3}{*}{\centering Mask R-CNN~\cite{he2017mask}
} 
 & 2048 &43.39  &31.05  &46.23  &74.05  &61.72  &49.89  &24.64  &19.94  &24.13  &27.15  &22.78  &16.24  &3.88  &3.29  &11.04  &34.62  &27.76  &29.51  \\
 & 4096 &43.13  &44.23  &37.30  &73.19  &70.08  &45.91  &27.31  &26.34  &26.69  &30.90  &30.16  &17.29  &17.58  &16.87  &9.64  &38.42  &37.53  &27.37  \\
 & 8192 &44.41  &44.80  &26.22  &74.22  &69.79  &49.46  &30.76  &29.33  &18.90  &34.62  &32.76  &20.39  &14.06  &11.11  &16.88  &39.61  &37.56  &26.37  \\
\midrule
\multirow{3}{*}{\centering\shortstack{Cascade R-CNN~\cite{8917599}}}
 & 2048 &59.05  &36.87  &43.71  &74.81  &62.59  &51.91  &29.68  &26.72  &30.80  &28.37  &25.07  &19.52  &8.20  &5.03  &12.71  &40.02  &31.26  &31.73  \\
 & 4096 &55.52  &52.94  &40.91  &73.09  &70.40  &50.74  &30.43  &29.53  &\textbf{35.99}  &33.22  &33.16  &18.64  &17.10  &17.85  &10.81  &41.87  &40.72  &31.42  \\
 & 8192 &50.88  &50.61  &34.15  &71.39  &67.88  &48.84  &31.14  &30.24  &20.74  &\textbf{36.45}  &35.37  &22.21  &24.42  &20.34  &\textbf{29.71}  &42.86  &40.89  &31.13  \\
\midrule
\multirow{3}{*}{\centering SOLOv2~\cite{wang2020solov2}} 
 & 2048 &56.40  &25.22  &49.49  &72.06  &33.31  &46.83  &30.56  &14.15  &27.64  &32.28  &14.32  &14.76  &10.91  &5.76  &19.19  &40.44  &18.55  &31.58  \\
 & 4096 &60.40  &32.98  &45.99  &72.35  &38.43  &50.82  &35.96  &22.12  &27.99  &31.00  &17.70  &14.93  &17.60  &12.84  &9.45  &43.46  &24.81  &29.84  \\
 & 8192 &58.08  &50.41  &30.11  &72.74  &49.93  &47.50  &37.52  &31.79  &13.97  &34.43  &25.75  &16.50  &14.38  &13.16  &13.85  &43.43  &34.21  &24.39  \\
\midrule
\multirow{3}{*}{\centering\shortstack{DiffusionInst~\cite{10447191}}}
 & 2048 &3.28  &0.00  &0.00  &0.02  &0.00  &0.00  &1.58  &0.00  &0.00  &2.13  &0.00  &0.00  &0.07  &0.00  &0.00  &1.42  &0.00  &0.00  \\
 & 4096 &30.41  &11.14  &12.52  &64.72  &23.37  &31.24  &17.18  &8.20  &6.17  &23.95  &11.06  &6.22  &2.49  &0.83  &0.24  &27.75  &10.92  &11.28  \\
  & 8192 & - & - & - & - & - & - & - & - & - & - & - & - & - & - & - & - & - & - \\
 \midrule
\multirow{3}{*}{\centering\shortstack{ASF-YOLO~\cite{kang2024asf}}}
 & 2048 &15.32  &10.68  &1.52  &54.87  &40.29  &9.90  &14.37  &9.73  &4.21  &21.79  &14.47  &5.06  &0.00  &0.00  &0.00  &21.27  &15.03  &4.14  \\
 & 4096 &19.78  &24.49  &6.02  &65.69  &57.08  &50.70  &17.26  &17.18  &8.58  &25.14  &22.91  &13.43  &0.00  &0.00  &0.00  &25.58  &24.33  &15.75  \\
 & 8192 &21.01  &22.00  &12.69  &58.63  &42.89  &42.98  &14.84  &11.90  &9.01  &22.84  &17.14  &13.60  &0.00  &0.00  &0.00  &23.46  &18.78  &15.66  \\
 \midrule
\multirow{3}{*}{\centering\shortstack{CelloType~\cite{pang2025cellotype}}}
 & 2048 &21.84  &3.78  &57.74  &58.00  &8.77  &62.90  &14.90  &2.85  &32.41  &20.77  &3.99  &18.03  &2.34  &0.85  &8.28  &23.57  &4.05  &35.87  \\
 & 4096 &21.59  &9.99  &\textbf{59.05}  &62.74  &19.56  &\textbf{64.97}  &15.93  &6.71  &26.73  &21.67  &8.01  &\textbf{23.91}  &3.56  &2.36  &20.16  &25.10  &9.33  &\textbf{38.96}  \\
 & 8192 & - & - & - & - & - & - & - & - & - & - & - & - & - & - & - & - & - & - \\
 \midrule
\multirow{3}{*}{\centering\shortstack{YOLOv12~\cite{tian2025yolov12}}}
 & 2048 &\textbf{80.62}  &\textbf{74.69}  &52.55  &70.34  &70.20  &45.21  &35.51  &37.46  &34.05  &33.49  &33.02  &11.56  &\textbf{39.08}  &\textbf{35.29}  &15.24  &51.81  &50.13  &31.72  \\
 & 4096 &71.09  &67.66  &50.54  &\textbf{77.26}  &\textbf{76.31}  &56.33  &\textbf{49.12}  &\textbf{48.78}  &28.22  &35.28  &\textbf{35.52}  &14.91  &28.42  &25.00  &11.99  &\textbf{52.23}  &\textbf{50.65}  &32.40  \\
 & 8192 &65.37  &61.49  &46.55  &67.33  &63.81  &38.70  &36.99  &37.50  &21.21  &26.72  &27.33  &7.00  &0.00  &0.00  &0.00  &39.28  &38.03  &22.69  \\
\bottomrule
\end{tabular}
\end{adjustbox}
\caption{Model performance comparison across five glomerular categories and three patch sizes (2048, 4096, 8192), using Dice, F1, and mAP metrics. - denotes configurations where computational constraints prevented evaluation.}
\label{tab:performance_comparison}
\end{table}

\newcommand{\down}{\ensuremath{\downarrow}}

\begin{table}[ht]
\centering
\begin{adjustbox}{width=\textwidth}
\renewcommand{\arraystretch}{1.2}
\begin{tabular}{c c ccc ccc ccc ccc ccc ccc}
\toprule
\multirow{2}{*}{\textbf{Method}} & \multirow{2}{*}{\textbf{Scale}} &
\multicolumn{3}{c}{\textbf{GGS}} & \multicolumn{3}{c}{\textbf{Viable Glom}} & \multicolumn{3}{c}{\textbf{Ischemic Glom}} &
\multicolumn{3}{c}{\textbf{SGS}} & \multicolumn{3}{c}{\textbf{Atubular Glom}} & \multicolumn{3}{c}{\textbf{Average}} \\
\cmidrule(lr){3-5} \cmidrule(lr){6-8} \cmidrule(lr){9-11} \cmidrule(lr){12-14} \cmidrule(lr){15-17} \cmidrule(lr){18-20}
 & & Dice\up & F1\up & mAP\up & Dice\up & F1\up & mAP\up & Dice\up & F1\up & mAP\up & Dice\up & F1\up & mAP\up & Dice\up & F1\up & mAP\up & Dice\up & F1\up & mAP\up \\
\midrule

\multirow{3}{*}{\centering \shortstack{Mask R-CNN~\cite{he2017mask}}} 
& 20$\times$ &43.13  &44.23  &37.30  &73.19  &70.08  &45.91  &27.31  &26.34  &26.69  &30.90  &30.16  &17.29  &17.58  &16.87  &9.64  &38.42  &37.53  &27.37  \\
& 10$\times$ &45.79 &48.73 &41.75 &75.15 &70.91 &59.32 &26.39 &25.17 &32.54 &27.94 &27.47 &21.61 &6.98 &9.72 &10.12 &36.45 &36.40 &33.07 \\
& 5$\times$ &51.63 &48.86 &38.54 &73.65 &68.91 &53.75 &25.96 &27.12 &26.78 &33.14 &33.16 &20.01 &18.34 &14.14 &13.30 &40.55 &38.44 &30.48 \\
\midrule

\multirow{3}{*}{\centering \shortstack{Cascade R-CNN~\cite{8917599}}} 
& 20$\times$ &55.52  &52.94  &40.91  &73.09  &70.40  &50.74  &30.43  &29.53  &35.99  &33.22  &33.16  &18.64  &17.10  &17.85  &10.81  &41.87  &40.72  &31.42  \\
& 10$\times$ &60.96 &57.38 &42.76 &72.86 &70.57 &47.78 &27.07 &26.82 &\textbf{36.30} &31.66 &30.52 &20.10 &22.80 &24.32 &13.76 &43.07 &41.92 &32.14 \\
& 5$\times$ &52.05 &50.76 &38.62 &74.92 &70.03 &53.43 &28.10 &28.14 &33.63 &32.36 &32.46 &17.95 &12.82 &12.24 &21.42 &40.05 &38.73 &33.01 \\
\midrule

\multirow{3}{*}{\centering \shortstack{SOLOv2~\cite{wang2020solov2}}} 
& 20$\times$   &60.40  &32.98  &45.99  &72.35  &38.43  &50.82  &35.96  &22.12  &27.99  &31.00  &17.70  &14.93  &17.60  &12.84  &9.45  &43.46  &24.81  &29.84 \\
& 10$\times$   & 57.67 & 30.03 & 46.17 & 73.22 & 37.09 & 51.93 & 34.28 & 18.06 & 33.22 & 30.96 & 17.55 & 14.48 & 16.23 & 10.98 & 19.77 & 42.47 & 22.74 & 33.11 \\
& 5$\times$    & 45.23 & 26.08 & 40.00 & 68.04 & 36.82 & 43.52 & 35.58 & 19.88 & 28.28 & \textbf{35.62} & 18.21 & 12.78 & 13.70 & 9.68 & 10.48 & 39.64 & 22.13 & 27.01 \\
\midrule

\multirow{3}{*}{\centering \shortstack{DiffusionInst~\cite{10447191}}} 
& 20$\times$ &30.41  &11.14  &12.52  &64.72  &23.37  &31.24  &17.18  &8.20  &6.17  &23.95  &11.06  &6.22  &2.49  &0.83  &0.24  &27.75  &10.92  &11.28 \\
& 10$\times$ &28.48 &12.67 &42.73 &71.24 &26.42 &55.64 &18.09 &9.71 &28.65 &25.41 &14.26 &14.02 &2.82 &2.39 &21.57 &29.21 &13.09 &32.52 \\
& 5$\times$ &27.41 &11.17 &44.49 &70.55 &22.21 &55.30 &17.43 &9.37 &34.02 &25.38 &11.46 &17.24 &3.41 &2.86 &21.96 &28.84 &11.42 &34.60 \\
\midrule

\multirow{3}{*}{\centering \shortstack{ASF-YOLO~\cite{kang2024asf}}} 
& 20$\times$ &19.78  &24.49  &6.02  &65.69  &57.08  &50.70  &17.26  &17.18  &8.58  &25.14  &22.91  &13.43  &0.00  &0.00  &0.00  &25.58  &24.33  &15.75 \\
& 10$\times$ &24.10 &33.61 &11.19 &65.05 &55.96 &56.18 &16.88 &16.71 &8.88 &24.81 &22.10 &12.77 &0.00 &0.00 &0.00 &26.17 &25.68 &17.80 \\
& 5$\times$ &22.31 &26.58 &7.84 &64.96 &53.40 &49.17 &16.78 &16.12 &8.15 &24.52 &21.68 &12.88 &0.00 &0.00 &0.00 &25.72 &23.56 &15.61 \\
\midrule

\multirow{3}{*}{\centering \shortstack{CelloType~\cite{pang2025cellotype}}} 
& 20$\times$ &21.59  &9.99  &\textbf{59.05}  &62.74  &19.56  &64.97  &15.93  &6.71  &26.73  &21.67  &8.01  &\textbf{23.91}  &3.56  &2.36  &20.16  &25.10  &9.33  &38.96  \\
& 10$\times$ &22.66 &9.52 &56.60 &63.57 &17.87 &\textbf{65.94} &16.12 &6.71 &28.35 &24.25 &10.68 &19.00 &3.32 &1.80 &30.81 &25.98 &9.32 &40.14 \\
& 5$\times$ &22.48 &9.34 &55.72 &65.70 &16.05 &65.06 &16.19 &7.37 &27.95 &22.67 &9.60 &22.27 &3.09 &2.46 &\textbf{34.86} &26.02 &8.96 &\textbf{41.17} \\
\midrule

\multirow{3}{*}{\centering \shortstack{YOLOv12~\cite{tian2025yolov12}}} 
& 20$\times$ &71.09  &67.66  &50.54  &77.26  &76.31  &56.33  &\textbf{49.12}  &\textbf{48.78}  &28.22  &35.28  &\textbf{35.52}  &14.91  &28.42  &25.00  &11.99  &52.23  &50.65  &32.40  \\
& 10$\times$ &65.51 &66.51 &54.07 &75.74 &73.46 &50.57 &40.59 &41.71 &25.73 &35.20 &34.69 &16.87 &21.81 &25.00 &12.73 &47.77 &48.27 &31.99 \\
& 5$\times$ &\textbf{73.51} &\textbf{68.66} &47.69 &\textbf{78.95} &\textbf{77.69} &60.10 &42.47 &42.86 &28.04 &33.06 &33.33 &15.66 &\textbf{34.26} &\textbf{37.84} &26.58 &\textbf{52.45} &\textbf{52.07} &35.62 \\
\bottomrule
\end{tabular}
\end{adjustbox}
\caption{Model performance across five glomerular categories at varying imaging magnifications (20$\times$, 10$\times$, 5$\times$). Metrics include Dice, F1, and mAP for each glomerular category.}
\label{tab:multi_magnification_results}
\end{table}

\begin{figure}[h]
\begin{center}
\includegraphics[width=0.95\textwidth]{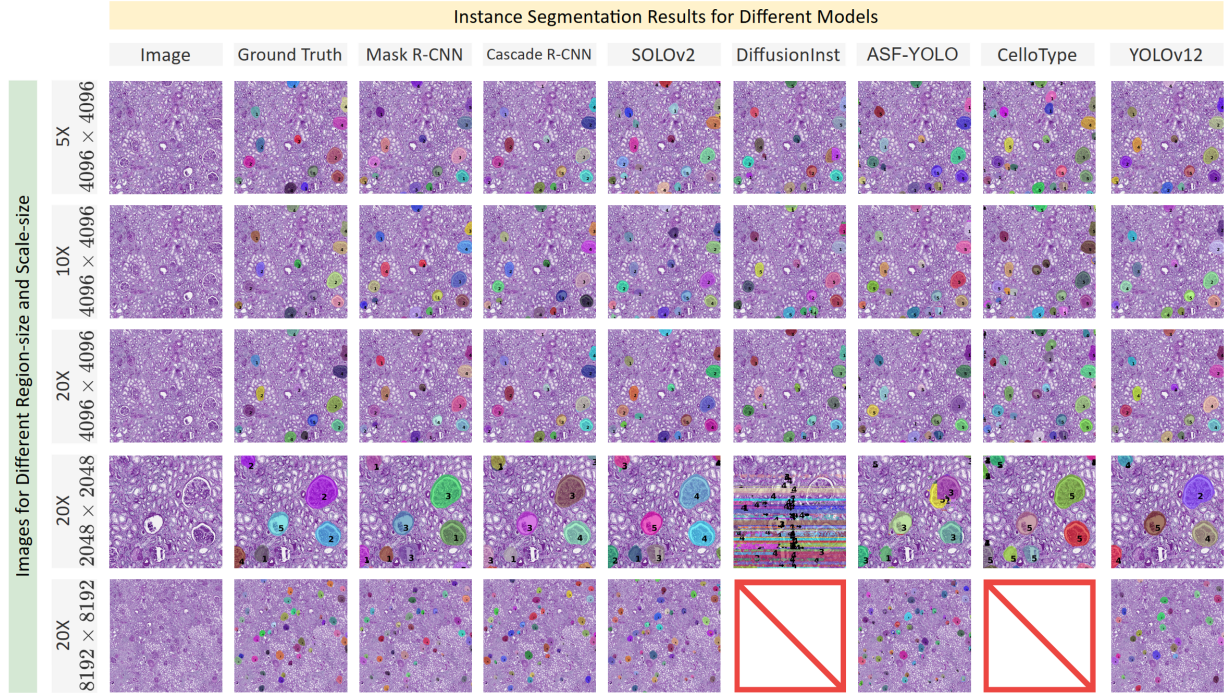}
\end{center}
\caption{Visualization of fine-grained diseased glomerular instance segmentation results across different region and scale sizes. The numbers 1 to 5 correspond respectively to GGS, Viable Glom, Ischemic Glom, SGS, Atubular Glom.} 
\label{fig2:visualize}
\end{figure}

\begin{figure}[h]
\begin{center}
\includegraphics[width=0.7\textwidth]{{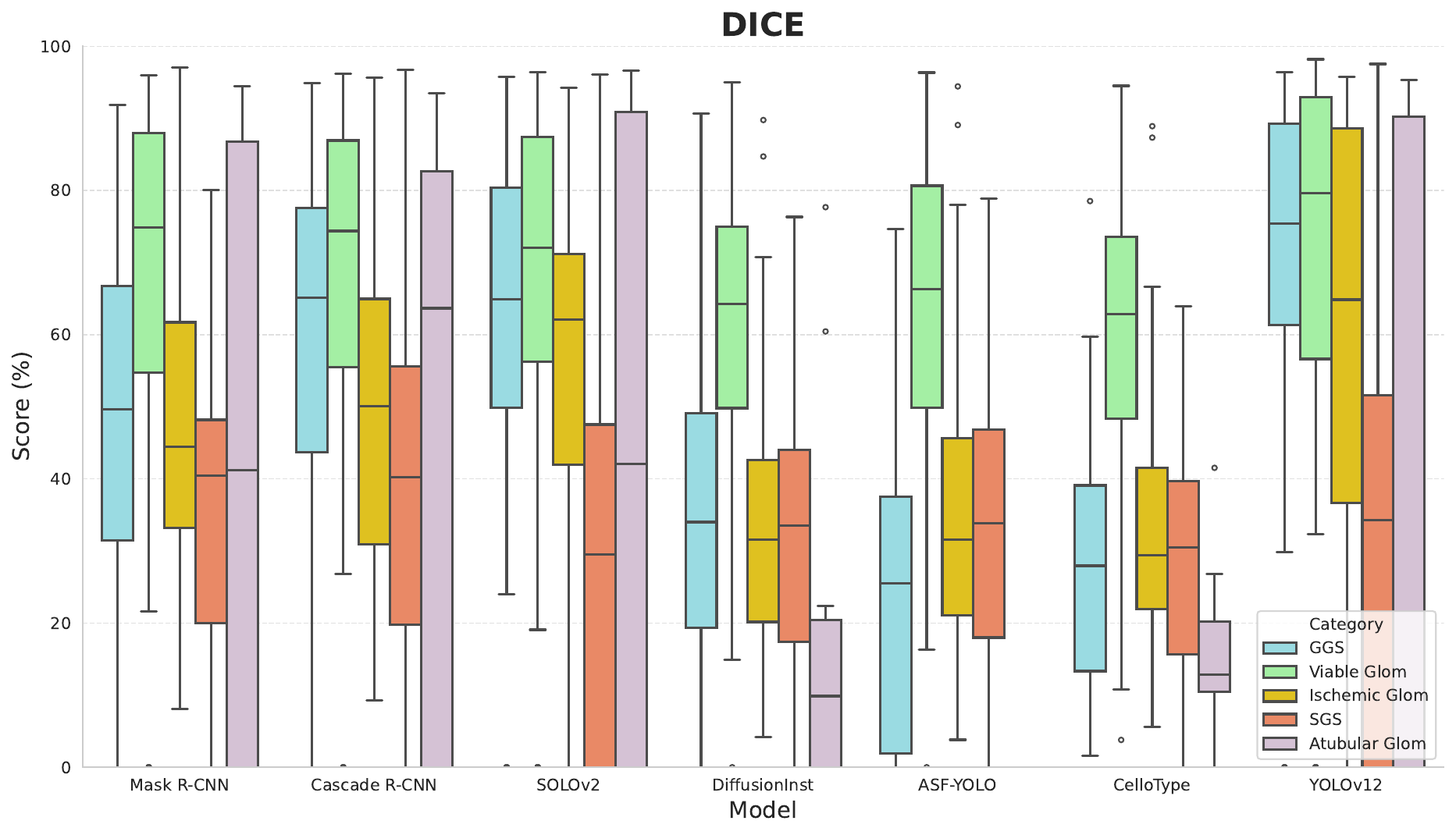}}
\end{center}
\caption{Boxplots of Dice scores for each glomerular category across seven segmentation methods using a patch size of 4096 pixels at 20$\times$. Each method includes five boxes corresponding to the five glomerular classes.}
\label{fig3:boxplot}
\end{figure}

\section{Results}

We evaluated seven representative models for glomerular instance segmentation across segmentation, classification, and detection tasks. Performance was assessed using Dice, F1, and mAP metrics.

\subsection{Overall Performance}

Table~\ref{tab:performance_comparison} presents the performance comparison across different patch sizes (2048, 4096, and 8192 pixels). Table~\ref{tab:multi_magnification_results} examines the effect of imaging magnification (20$\times$, 10$\times$, and 5$\times$) on segmentation performance, where higher magnifications provide greater image clarity while maintaining the same field of view.

\noindent \textbf{Segmentation:} YOLOv12 achieved the strongest segmentation accuracy, with an average Dice of 52.23\% and F1 of 50.65\% at the 4096-pixel patch size. It also delivered the best Dice score (39.08\%) for Atubular Glom, the rarest category, highlighting its robustness to class imbalance. Cascade R-CNN and SOLOv2 produced balanced but mid-range results, whereas DiffusionInst and ASF-YOLO underperformed with unstable scores. CelloType consistently achieved the highest mAP values, reflecting strong localization despite weaker mask quality. 

Figure~\ref{fig3:boxplot} demonstrates marked differences in the distribution of Dice scores across glomerular categories. Viable Glom and GGS show relatively high medians with compact interquartile ranges, reflecting both strong central performance and consistency across patches. By contrast, Ischemic and SGS Glom display lower medians and noticeably wider score distributions, indicating greater heterogeneity and less stable segmentation quality. Atubular Glom exhibit the broadest spread, with performance ranging from near-random to high accuracy, underscoring the difficulty of consistently delineating this rare and morphologically diverse subtype. These score distributions reveal that class imbalance and structural variability are key drivers of performance instability in diseased glomerular segmentation.

\noindent \textbf{Classification:} Category-specific sensitivity varied across methods. Categories with clear morphological patterns, such as GGS and Atubular Glom, were better detected at smaller patch sizes. Ischemic Glom remained the most challenging across all models, suggesting the need for stronger class-specific feature learning.

\noindent \textbf{Detection:} While CelloType excelled in AP-based localization, YOLOv12 provided the best trade-off between detection and segmentation, making it the most reliable candidate for clinical translation.

\subsection{Multi-Region Performance}

The result indicated that patch size significantly influenced outcomes. The 4096-pixel configuration consistently offered the best balance, capturing sufficient context without introducing noise or excessive computational burden. Smaller regions (2048) improved detection for distinct classes like GGS and Atubular Glom, while larger regions (8192) led to performance declines and instability. DiffusionInst and CelloType could not be evaluated at 8192 due to computational constraints, which could impact their practical applicability in clinical settings with limited computational infrastructure.

\subsection{Multi-Scale Performance}

Magnification analysis revealed that maximum clarity does not always yield optimal results. At 20$\times$, fine structural details were captured effectively, but 10$\times$ often produced better balance between Dice and mAP, particularly for Mask R-CNN and Cascade R-CNN. Surprisingly, YOLOv12 achieved its best overall performance at 5$\times$, suggesting that reduced clarity may suppress irrelevant fine details, enabling more robust generalization. These findings underscore that optimal performance depends on aligning resolution with model architecture, rather than uniformly maximizing image clarity.

\subsection{Discussion and Future Work}

We systematically evaluated glomerular instance segmentation across multiple patch sizes and magnifications, revealing key determinants of performance. Across methods, the 4096 patch size consistently provided the most balanced results, optimizing both contextual coverage and computational efficiency, with YOLOv12 achieving the highest Dice (52.23\%) and F1 (50.65\%) scores. Notably, CelloType consistently achieved the highest mAP values despite lower Dice and F1 scores, highlighting its strength in object localization, which future methods could leverage to improve detection precision.    Magnification analysis showed that maximal clarity was not always optimal: 20$\times$ magnification improved fine-detail recognition, but some methods achieved peak overall performance at 5$\times$, suggesting that moderate image smoothing can enhance generalization and reduce overfitting. Taken together, these results highlight the importance of selecting intermediate patch sizes and appropriate magnifications, and suggest future work on adaptive scale-aware architectures and patching strategies that combine robust localization with accurate mask prediction to further improve lesion-level glomerular segmentation.

\section{New or Breakthrough Work to be Presented}

This study introduces M$^3$-GloDets, the first systematic benchmarking framework dedicated to diseased glomerular detection and segmentation. In contrast to earlier efforts that concentrated mainly on classification or semantic segmentation of normal glomeruli and broad lesion categories, our framework advances the field by enabling rigorous instance-level evaluation across multiple diseased subtypes. M$^3$-GloDets establishes a unified protocol to compare diverse model architectures under varying region sizes, magnifications, and class distributions, reflecting the complexity of digital renal pathology.  This work fills a critical methodological gap, setting new standards for evaluating glomerular detection pipelines and providing actionable insights for algorithm design, dataset construction, and clinical translation in renal pathology.

\section{Conclusion} 

In this study, we systematically evaluated multiple instance segmentation frameworks for glomerular structures across different patch sizes and magnifications. YOLOv12 achieved the highest overall Dice and F1 scores, including robust detection of rare Atubular Glom, while CelloType consistently attained the highest mAP, highlighting the value of localization-focused strategies. Intermediate patch sizes (4096 pixels) provided the best balance between context and efficiency, and moderate magnifications enhanced generalization by reducing overfitting. These findings emphasize that careful selection of resolution, magnification, and architecture is critical for reliable lesion-level segmentation, and suggest future work on adaptive scale-aware designs and methods that combine strong localization with accurate mask prediction to further advance glomerular pathology analysis.

\section{ACKNOWLEDGMENTS} 
This research was supported by the WCM Radiology AIMI Fellowship and WCM CTSC 2026 Pilot Award.

\bibliography{main} 
\bibliographystyle{spiebib} 

\end{document}